\documentclass[conference]{IEEEtran}
\IEEEoverridecommandlockouts
\usepackage{cite}
\usepackage{verbatim}
\usepackage{amsmath,amssymb,amsfonts}
\usepackage{algorithmic}
\usepackage{algorithm}
\usepackage{graphicx}
\usepackage{booktabs}
\usepackage{color,soul}
\usepackage{textcomp}
\usepackage{environ}
\usepackage{color}
\usepackage{xcolor}
\usepackage{multirow}
\usepackage{multicol}
\usepackage{subcaption}
\newcommand{\Lrec}{\mathcal{L}_{\mathrm{rec}}}
\newcommand{\Lcls}{\mathcal{L}_{\mathrm{cls}}}
\newcommand{\Lkl}{\mathcal{L}_{\mathrm{KL}}}
\newcommand{\Ltot}{\mathcal{L}_{\mathrm{total}}}
\newcommand{\SSIM}{\mathrm{SSIM}}

\newcommand{\bmu}{\boldsymbol{\mu}}
\newcommand{\bsig}{\boldsymbol{\sigma}}
\newcommand{\beps}{\boldsymbol{\varepsilon}}

\def\BibTeX{{\rm B\kern-.05em{\sc i\kern-.025em b}\kern-.08em
    T\kern-.1667em\lower.7ex\hbox{E}\kern-.125emX}}

\begin{document}

\title{\fontsize{18.75}{18}\selectfont{A VAE-Driven Multi-Task Satellite-Aided Semantic Communication Framework for 6G-Enabled Connected Autonomous Vehicles}
\vspace{-3.5 mm}
}

\author{
\IEEEauthorblockN{
S. M. Abtahiul Alam\textsuperscript{1},
Niloy Das\textsuperscript{1},
Apurba Adhikary\textsuperscript{1},
Yu Qiao\textsuperscript{2},
Zhu Han\textsuperscript{3}, and
Choong Seon Hong\textsuperscript{4}
}
\IEEEauthorblockA{
\textsuperscript{1}\textit{Department of Information and Communication Engineering,}
\textit{Noakhali Science and Technology University, Bangladesh}\\
\textsuperscript{2}\textit{Department of Artificial Intelligence,}
\textit{Kyung Hee University, Yongin-si 17104, Republic of Korea}\\
\textsuperscript{3}\textit{Department of Electrical and Computer Engineering,}
\textit{University of Houston, Houston, TX 77004-4005, USA}\\
\textsuperscript{4}\textit{Department of Computer Science and Engineering,}
\textit{Kyung Hee University, Yongin-si 17104, Republic of Korea}\\
{\footnotesize E-mail: abtahi1116@student.nstu.edu.bd, dasnil684@gmail.com, apurba@nstu.edu.bd, qiaoyu@khu.ac.kr, zhan2@uh.edu, cshong@khu.ac.kr}
}
\vspace{-10 mm}
}
\maketitle

\begin{abstract}
The development of smart transportation systems and the introduction of 6G wireless communication technologies have significantly changed vehicle network topologies. Future connected autonomous vehicle (CAV) networks require bandwidth-efficient, reliable, and low-latency communication for safety-critical applications such as traffic sign recognition and decision-making. Conventional communication systems transmit raw data regardless of task relevance, which is inefficient in resource-constrained satellite channels where uplink bandwidth is scarce and propagation losses are large. Semantic communication addresses this limitation by transmitting task-relevant information instead of full signal representations. It extracts and conveys essential semantic features and leverages deep learning to optimize task performance at the receiver. Therefore, we present a Variational Autoencoder (VAE)-based multi-task semantic communication framework for satellite-assisted autonomous driving. Unlike deterministic autoencoder-based methods, the proposed model uses probabilistic latent representations for more robust and efficient encoding. The learned features are transmitted over noisy wireless channels to perform traffic sign reconstruction and classification. The framework is trained end-to-end to jointly optimize both tasks. Results show that the proposed approach achieves significant bandwidth reduction of up to 87.23\% to 98.17\% while maintaining stable performance across varying signal-to-noise ratio conditions.
\end{abstract}



\section{Introduction}
\label{sec:intro}


Autonomous vehicles in rural or disaster areas increasingly depend on satellite communication to share vital data. When a sender vehicle detects a traffic sign, it must transmit not just the image but actionable semantic information for the receiver vehicle to identify the sign in real time. This procedure is difficult because of the limited bandwidth of satellites, propagation losses, and channel problems like fading and noise. Conventional methods that separate compression, modulation, and inference are inefficient, as they transmit pixel-level data without addressing the receiver’s needs.  In semantic communication, one learns compact representations that maintain the information related to the tasks. The foundational work on joint source and channel coding~(JSCC) demonstrated that end-to-end training of transmission and reception implicitly takes into account channel effects. Most of the research efforts have been dedicated to single-task problems, e.g., reconstruction and classification ~\cite{bourtsoulatze2019deep, xie2020lite, raha2022goal, raha2023artificial}. Few studies exist on joint task learning, and those that do typically use mean squared error (MSE) for reconstruction ~\cite{eldeeb2024multi}.

While MSE is convenient, it has limitations in perceptual quality. Decoders minimizing pixel-wise squared error often produce blurred outputs, averaging over plausible reconstructions instead of choosing a definitive option. This is particularly problematic for traffic sign images, where precise geometry is crucial for distinguishing similar categories. Blurred boundaries between characters or shapes reduce both the image quality and the discriminative signal necessary for effective classification, as both the decoder and classifier rely on the same latent vector. 

This work proposes a VAE-based multi-task semantic communication framework for satellite-assisted vehicular networks, in which MSE-only reconstruction is replaced by a composite perceptual loss combining MSE, structural similarity (SSIM), and image-gradient terms. A two-branch decoder with a residual refinement sub-network recovers detail suppressed by the communication bottleneck, and a deterministic inference strategy, substituting the posterior mean for a stochastic sample at test time, stabilizes reconstruction quality without sacrificing the benefits of probabilistic training. The main contributions are as follows:

\begin{itemize}\setlength\itemsep{1 em}

    \item We propose a satellite-assisted semantic communication framework for connected autonomous vehicles that enhances efficiency and robustness by using task-oriented feature encoding instead of raw pixel-level transmission.
    
    \item We employ a probabilistic latent representation with a dimension-scaled Kullback-Leibler~(KL) weight to enhance robustness under high noise levels in satellite channels and to prevent posterior collapse at small codeword dimensions.
    
    \item We introduce a unified multi-task framework that jointly optimizes reconstruction and classification, incorporating a two-branch decoder with a residual refinement sub-network that recovers fine-grained edge detail suppressed by the communication bottleneck.

    \item We test our approach against conventional methods on different traffic sign datasets under different channel conditions, including shadowed and clear-sky satellite propagation.

\end{itemize}

The rest of the paper is organized as follows: \ref{sec:related} reviews related works in semantic communication and deep learning-based communication systems. \ref{sec:sysmodel} presents the system model. \ref{sec:method} describes the proposed methodology. \ref{sec:exp} details the experimental setup. \ref{sec:results} presents the results and analysis, and finally, \ref{sec:conclusion} concludes the paper.

\section{Related Work}
\label{sec:related}

Deep joint source–channel coding (DeepJSCC)~\cite{bourtsoulatze2019deep} replaces the conventional separation of source and channel coding with end-to-end learning over noisy channels, enabling graceful degradation under channel impairments. Extensions have explored feedback mechanisms and advanced architectures such as transformers and diffusion-based receivers, although their high computational cost and latency limit deployment in real-time or resource-constrained scenarios, where lightweight convolutional models remain more practical. Moreover, most prior work focuses on terrestrial or Additive White Gaussian Noise~(AWGN) channels, while satellite-specific challenges such as large path loss, Doppler shifts in LEO systems, and two-hop relay geometry remain underexplored, motivating this study. Recent task-oriented semantic communication methods~\cite{eldeeb2024multi} demonstrate shared latent spaces for joint reconstruction and classification, but reliance on pixel-wise losses obscures the contribution of latent modeling versus architectural design; in contrast, this work introduces a composite perceptual loss and a matched deterministic baseline to isolate the effects of loss design and stochastic latent representations. Furthermore, while MSE-based objectives often yield overly smooth reconstructions, perceptual and structural metrics such as SSIM~\cite{wang2004image} and feature-based losses~\cite{johnson2016perceptual}, together with gradient-based constraints~\cite{ledig2017photo}, better preserve visual fidelity, motivating our composite loss formulation.

\section{System Model and Problem Formulation}
\label{sec:sysmodel}
\subsection{Transmission Scenario}

We consider a satellite-assisted vehicular communication scenario where a sender autonomous vehicle captures a traffic sign image and transmits a compressed semantic codeword to a satellite relay, which forwards it to a receiver vehicle. The end-to-end uplink–downlink path is modeled as a single equivalent flat-fading AWGN channel using the standard two-hop abstraction in satellite systems~\cite{eldeeb2024multi}, valid under decode- or amplify-and-forward relaying with an equivalent received signal-to-noise ratio~(SNR) denoted as $\gamma$. Although satellite links often exhibit strong line-of-sight conditions, we adopt Rayleigh fading as a conservative worst-case model to account for urban blockage, foliage shadowing, and low-elevation passes, ensuring robust performance evaluation. The transmitter maps an image $\mathbf{x} \in [0,1]^{H \times W}$ with class label $c \in \mathcal{Y} = \{0,\ldots,C-1\}$ to a latent codeword $\mathbf{z} \in \mathbb{R}^d$, which is transmitted over the channel. The received signal $\mathbf{y} \in \mathbb{R}^d$ is then used to jointly reconstruct the image $\hat{\mathbf{x}}$ and predict the label $\hat{c}$. A single fading coefficient $h$ is assumed constant across all $d$ dimensions within one transmission interval, corresponding to a narrowband channel.

\subsection{Problem Formulation}

Let $f_\phi: \mathbb{R}^{H \times W} \rightarrow \mathbb{R}^d$ denote the encoder with parameters $\phi$, and let $g_\theta: \mathbb{R}^d \rightarrow \mathbb{R}^{H \times W}$ and $q_\psi: \mathbb{R}^d \rightarrow \mathbb{R}^C$ denote the reconstruction and classification networks at the receiver, respectively. The system is trained in an end-to-end manner by solving the following constrained optimization problem:
\begin{equation}
\begin{aligned}
\min_{\phi,\,\theta,\,\psi}\quad 
\mathbb{E}_{\mathbf{x},c,\mathbf{y}}\Big[
&\alpha\,\mathcal{L}_{\mathrm{cls}}\!\left(q_\psi(\mathbf{y}), c\right)
+ \beta\,\mathcal{L}_{\mathrm{rec}}\!\left(g_\theta(\mathbf{y}), \mathbf{x}\right) \\
&+ \gamma\,\mathcal{L}_{\mathrm{kl}}\!\left(f_\phi(\mathbf{x})\right)
\Big],
\end{aligned}
\label{eq:objective}
\end{equation}
\vspace{-6mm}
\begin{align}
\text{s.t.}\quad 
&\mathbb{E}\!\left[\|\mathbf{z}\|_2^2\right] \leq P, \tag{1a} \label{eq:c1} \\
&d \leq d_{\max}, \tag{1b} \label{eq:c2} \\
&\hat{\mathbf{x}} = g_\theta(\mathbf{y}) \in [0,1]^{H \times W}, \tag{1c} \label{eq:c3} \\
&\mathbf{y} = h\,\tilde{\mathbf{z}} + \mathbf{n}, \quad 
h \sim \mathrm{Rayleigh},\;
\mathbf{n} \sim \mathcal{N}(\mathbf{0}, \sigma_n^2 \mathbf{I}), \tag{1d} \label{eq:c4}
\end{align}
where $\mathbf{z} = f_\phi(\mathbf{x})$ denotes the encoded latent representation and $\tilde{\mathbf{z}}$ is its power-normalized version. Constraint~\eqref{eq:c1} enforces a transmit power budget $P$, while~\eqref{eq:c2} limits the codeword dimension to $d_{\max}$ based on available bandwidth. Constraint~\eqref{eq:c3} ensures valid pixel reconstruction, and~\eqref{eq:c4} defines the satellite channel model, where $\sigma_n^2 = 1/(2\,\mathrm{\gamma}_{\mathrm{lin}})$. The objective jointly optimizes classification accuracy, reconstruction fidelity, and latent regularization, enabling robust end-to-end semantic communication over noisy wireless channels. The regularization term $\mathcal{L}_{\mathrm{kl}}(\cdot)$ is included only in the probabilistic VAE formulation and is omitted for the deterministic baseline. In practice, the power constraint in~\eqref{eq:c1} is enforced by normalizing $\mathbf{z}$ to unit RMS amplitude before transmission, as detailed in the methodology Section~\ref{sec:method}.

\section{Proposed Framework}
\label{sec:method}


\begin{figure*}[!t]
    \centering
    \includegraphics[width=1\linewidth]{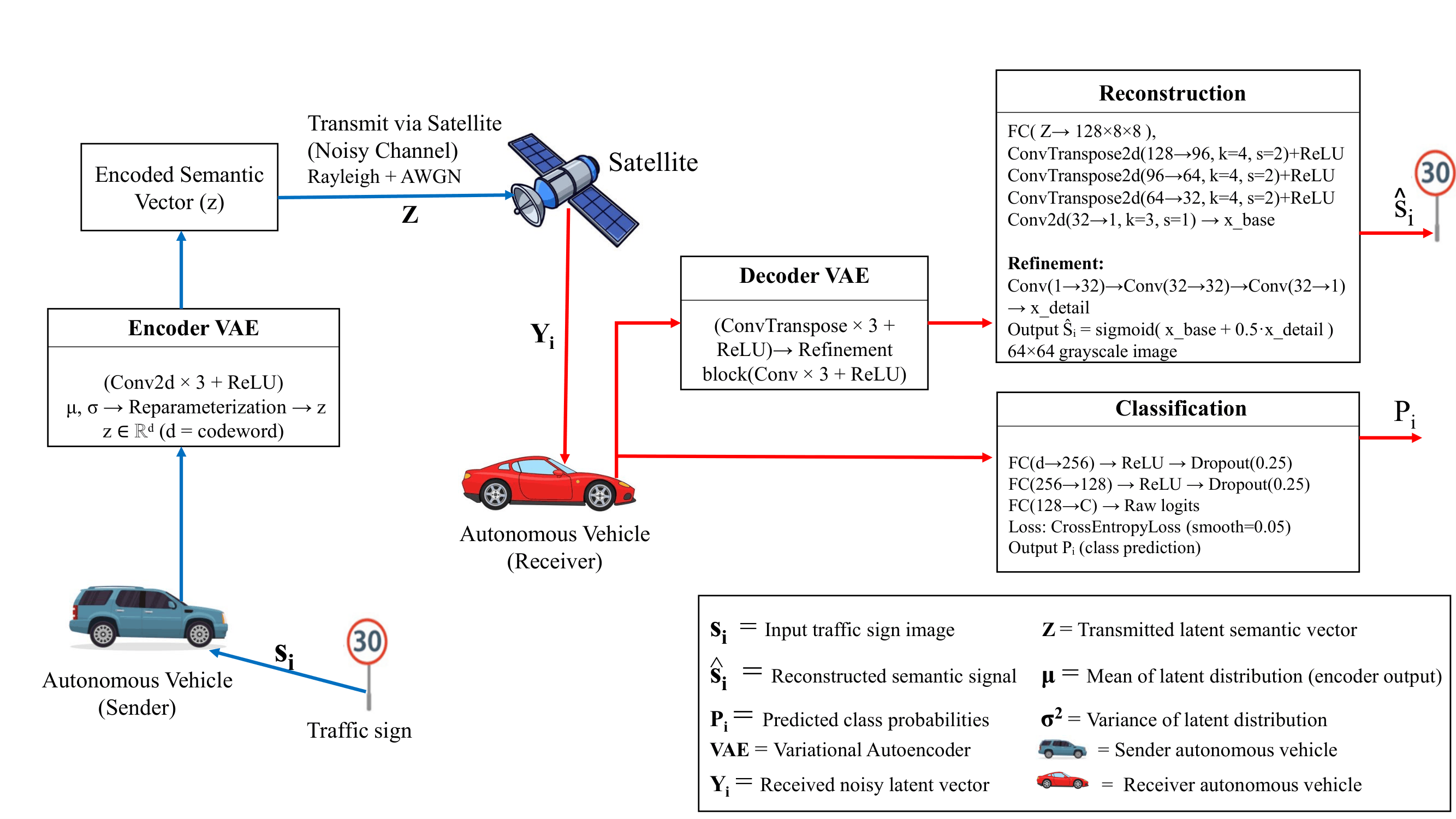}
    \caption{VAE-based multi-task satellite-aided semantic communication framework for 6G autonomous vehicles, featuring a refinement branch for better reconstruction and a parallel classifier for traffic sign recognition in noisy channels.}
    \label{fig:framework}
    \vspace{-6mm}
\end{figure*}

\subsection{Encoder Architecture}

\begin{figure}
    \centering
    \includegraphics[width=1\linewidth]{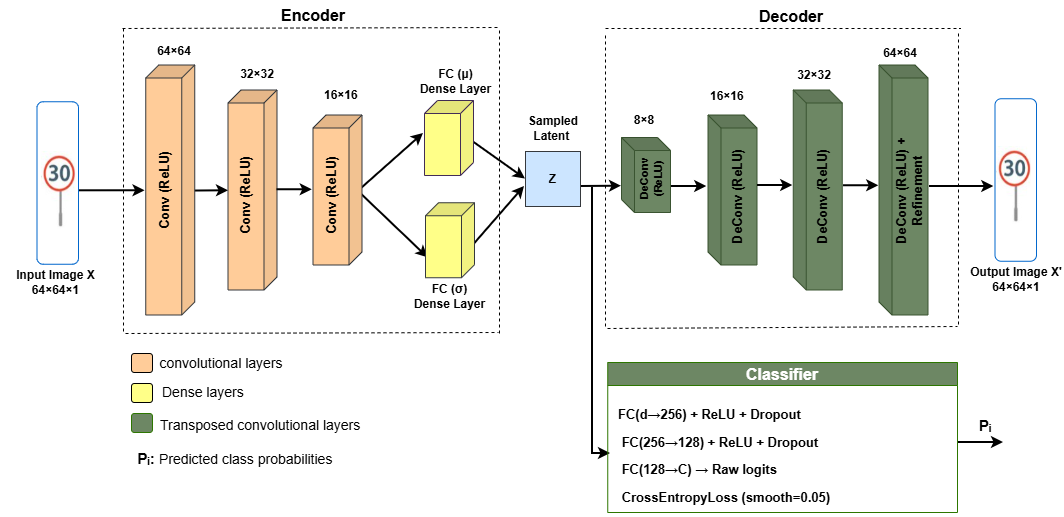}
    \caption{Proposed encoder–decoder architecture of VAE with latent sampling and a refinement branch.}
    \label{fig:vae_arc}
    \vspace{-6mm}
\end{figure}

The encoder \( f_\phi \) comprises three strided convolutional blocks that progressively reduce the spatial resolution, as shown in Fig.~\ref{fig:vae_arc}. Each block processes an input spatial map and produces a downsampled output. The convolution blocks are denoted as \( \text{C}(k,s,p) \), where \( k \) is the kernel size, \( s \) is the stride, and \( p \) is the padding. The resulting feature volume is flattened into a vector \( \mathbf{h} \in \mathbb{R}^{8192} \), which is then passed through two separate fully connected layers to yield the posterior mean and log-variance:
\begin{equation}
\bmu = \mathbf{W}_\mu \mathbf{h} + \mathbf{b}_\mu, \quad
\log\bsig^2 = \mathbf{W}_v \mathbf{h} + \mathbf{b}_v,
\label{eq:encoder_out}
\end{equation}
each producing a $d$-dimensional vector. These two branches correspond to fully connected dense layers, specifically FC($\mu$) and FC($\sigma$) layers, as shown in Fig.~\ref{fig:vae_arc}.


\subsection{Reparameterization and Channel Model}

During training, the latent vector is drawn via the reparameterization trick~\cite{kingma2013auto}:

\begin{equation}
\mathbf{z} = \bmu + \beps \odot \bsig, \quad \beps \sim \mathcal{N}(\mathbf{0},\mathbf{I}),
\label{eq:reparam}
\end{equation}
where $\bsig = \exp\!\left(0.5\,\log\bsig^2\right)$. At inference, we substitute $\mathbf{z} = \bmu$ for deterministic evaluation, eliminating stochastic reconstruction variance and consistently improving perceptual quality on the test set. Before transmission, $\mathbf{z}$ is normalized by its root-mean-square amplitude to decouple signal power from input scale:
\begin{equation}
\tilde{\mathbf{z}} = \frac{\mathbf{z}}{\displaystyle\sqrt{\frac{1}{d}\sum_{i=1}^d z_i^2}}.
\label{eq:power_norm}
\end{equation}

The channel introduces Rayleigh fading and AWGN:
\begin{equation}
\mathbf{y} = h\,\tilde{\mathbf{z}} + \mathbf{n},
\label{eq:channel}
\end{equation}
where the fading coefficient is drawn element-wise as
\begin{equation}
h = \sqrt{\frac{h_R^2 + h_I^2}{2}}, \quad h_R,h_I \sim \mathcal{N}(0,1),
\label{eq:rayleigh}
\end{equation}
and the noise is $\mathbf{n} \sim \mathcal{N}(\mathbf{0},\sigma_n^2\mathbf{I})$ with standard deviation
\begin{equation}
\sigma_n = \frac{1}{\sqrt{2\,\gamma_{\mathrm{lin}}}},\quad
\gamma_{\mathrm{lin}} = 10^{\gamma_{\mathrm{dB}}/10}.
\label{eq:noise_std}
\end{equation}
No channel state information is assumed at the transmitter; both the decoder and classifier learn to operate within the faded, noisy signal space defined by the training channel distribution. The received signal $\mathbf{y}$ is rescaled by the original RMS power before being passed to the decoder and classifier.

\subsection{Decoder with Refinement Branch}

The decoder transforms the vector \(\mathbf{y}\) into image space through a fully connected projection and four transposed convolutional blocks, following the encoder's downsampling hierarchy. The first three blocks enhance spatial resolution while decreasing channel depth, while the fourth block keeps a \(64 \times 64\) resolution and has a refinement branch. 
Transposed convolution is defined as \(\text{CT}(k,s,p)\) with kernel size \(k\), stride \(s\), and padding \(p\), followed by a ReLU activation. In the final stage, a base map \(\mathbf{x}_\mathrm{base} \in \mathbb{R}^{64 \times 64}\) is generated using a \(3 \times 3\) convolution (from \(32\) to \(1\) channel), alongside a three-layer refinement sub-network.
\begin{equation}
\begin{aligned}
\mathbf{x}_\mathrm{detail} \in
&\ \mathrm{Conv}_{3\times3}^{1\to32} \rightarrow \mathrm{ReLU} \rightarrow \\
&\ \mathrm{Conv}_{3\times3}^{32\to32} \rightarrow \mathrm{ReLU} \rightarrow \mathrm{Conv}_{3\times3}^{32\to1}
\end{aligned}
\label{eq:refine}
\end{equation}
and the two branches are fused as:
\begin{equation}
\hat{\mathbf{x}} = \sigma\!\left(\mathbf{x}_\mathrm{base}
+ 0.5\,\mathbf{x}_\mathrm{detail}\right),
\label{eq:decoder_out}
\end{equation}
where $\sigma(\cdot)$ is the sigmoid function.

\subsection{Classifier Design}

We use a classifier $q_\psi$ to map the received signal $\mathbf{y}$ to class logits using two fully-connected layers with dropout ($p=0.25$), forming a compact nonlinear decision head:
\begin{equation}
\mathbb{R}^d \rightarrow 256 \rightarrow 128 \rightarrow C.
\label{eq:classifier}
\end{equation}

\subsection{Composite Reconstruction Loss}
\label{subsec:loss_rec}

To improve perceptual fidelity beyond pixel-wise supervision, we design a composite reconstruction loss combining intensity, structural, and edge-aware objectives:
\begin{equation}
\Lrec =
0.4\,\mathcal{L}_1
+ 0.1\,\mathcal{L}_\mathrm{MSE}
+ 0.3\,(1 - \SSIM)
+ 0.2\,\mathcal{L}_\mathrm{edge}.
\label{eq:lrec}
\end{equation}
Where $\mathcal{L}_1$ preserves sharpness, MSE stabilizes optimization, SSIM~\cite{wang2004image} enforces structural similarity, and the edge loss preserves boundary information critical for traffic sign semantics.

\begin{algorithm}[!t]
\caption{Satellite-Assisted Semantic Transmission and Inference}
\label{alg:inference}
\begin{algorithmic}[1]
\REQUIRE Trained encoder $f_\phi$, decoder $g_\theta$,
         classifier $q_\psi$; test image $\mathbf{x}$;
         channel SNR$_{\mathrm{test}}$
\ENSURE Reconstructed image $\hat{\mathbf{x}}$;
        predicted label $\hat{c}$
\STATE $(\bmu, \log\bsig^2) \leftarrow f_\phi(\mathbf{x})$
       \COMMENT{Encode at sender vehicle}
\STATE $\tilde{\mathbf{z}} \leftarrow \bmu \,/\,
       \sqrt{\tfrac{1}{d}\sum_i \mu_i^2}$
       \COMMENT{Power normalise; use mean (no sampling)}
\STATE /* \textit{Uplink: sender vehicle $\rightarrow$ satellite} */
\STATE $h_{\uparrow} \sim \mathrm{Rayleigh}(1)$;\quad
       $\mathbf{n}_{\uparrow} \sim
       \mathcal{N}(\mathbf{0},\sigma_n^2\mathbf{I})$
\STATE $\mathbf{r} \leftarrow h_{\uparrow}\,\tilde{\mathbf{z}}
       + \mathbf{n}_{\uparrow}$
       \COMMENT{Received signal at satellite}
\STATE /* \textit{Downlink: satellite $\rightarrow$ receiver vehicles} */
\STATE $h_{\downarrow} \sim \mathrm{Rayleigh}(1)$;\quad
       $\mathbf{n}_{\downarrow} \sim
       \mathcal{N}(\mathbf{0},\sigma_n^2\mathbf{I})$
\STATE $\mathbf{y} \leftarrow h_{\downarrow}\,\mathbf{r}
       + \mathbf{n}_{\downarrow}$
       \COMMENT{Received signal at receiver vehicle}
\STATE $\hat{\mathbf{x}} \leftarrow g_\theta(\mathbf{y})$
       \COMMENT{Reconstruct image}
\STATE $\hat{c} \leftarrow \arg\max\; q_\psi(\mathbf{y})$
       \COMMENT{Classify traffic sign} 
\RETURN $\hat{\mathbf{x}},\;\hat{c}$
\end{algorithmic}
\end{algorithm}
\setlength{\textfloatsep}{0pt}
Algorithm~\ref{alg:inference} summarizes the inference procedure at deployment, explicitly showing the two-hop satellite relay channel
through which the semantic codeword passes before reconstruction and
classification at the receiver vehicle.

\subsection{Training Objective}

The classification loss is defined using cross-entropy with label smoothing:
\begin{equation}
\Lcls = -\sum_{c=0}^{C-1}\tilde{y}_c\log\hat{p}_c.
\label{eq:cls}
\end{equation}
The latent regularization term is given by the KL divergence:
\begin{equation}
\Lkl = \frac{1}{d}\left(
-\frac{1}{2}\sum_{i=1}^d
(1+\log\sigma_i^2-\mu_i^2-\sigma_i^2)
\right),
\label{eq:kl}
\end{equation}
with adaptive scaling:
\begin{equation}
\gamma(d) = \gamma_0\cdot\frac{16}{d}, \quad \gamma_0=0.01.
\label{eq:gamma_scale}
\end{equation}
The overall optimization objective is:
\begin{equation}
\Ltot = \alpha\,\Lcls + \beta\,\Lrec + \gamma(d)\,\Lkl,
\quad \alpha=0.8,\;\beta=1.8.
\label{eq:total}
\end{equation}

\subsection{Simulation Setup}
\label{sec:exp}

\subsubsection{Dataset}

We evaluate the proposed framework on two benchmark traffic sign datasets: the Chinese Traffic Sign dataset~\cite{yang2015towards} and the German Traffic Sign Recognition Benchmark (GTSRB)~\cite{stallkamp2011german}. Both datasets consist of grayscale images resized to $64 \times 64$ pixels and cover multiple fine-grained traffic sign categories. To ensure sufficient class representation and training stability, we retain only categories with a minimum number of samples per class. After filtering, 30 classes are selected from the Chinese dataset and 35 classes from GTSRB based on their sample frequency. The dataset splits and training configurations are summarized in Table~\ref{tab:sim_params}.

\begin{table}[!t]
\centering
\caption{Dataset and Training Configuration}
\label{tab:sim_params}
\setlength{\tabcolsep}{12pt}
\renewcommand{\arraystretch}{1.1}
\begin{tabular}{lcc}
\toprule
\textbf{Parameter} & \textbf{Chinese Dataset} & \textbf{German Dataset} \\
\midrule
Training samples   & 2964  & 12270 \\
Validation samples  & 524   & 2630  \\
Test samples       & 1756  & 10110 \\
\midrule
Image resolution   & $64 \times 64$ & $64 \times 64$ \\
Batch size         & 32    & 32 \\
Training epochs    & 100   & 100 \\
Optimizer          & AdamW & AdamW \\
Learning rate      & $5 \times 10^{-4}$ & $5 \times 10^{-4}$ \\
Weight decay       & $1 \times 10^{-4}$ & $1 \times 10^{-4}$ \\
Selected classes    & 30    & 35 \\
\bottomrule
\end{tabular}
\vspace{-3mm}
\end{table}

\subsubsection{Baselines}

We compare the proposed method against two representative baselines:

\begin{enumerate}
    \item Convolutional Autoencoder (AE): A deterministic semantic communication model that employs the same encoder--decoder architecture as the proposed method but excludes stochastic latent sampling and KL regularization. The training objective includes reconstruction and classification losses only.
    \item Quadrature Amplitude Modulation (QAM): A conventional digital communication system in which images are quantized and transmitted using 16-QAM modulation over a Rayleigh fading channel. At the receiver, a linear equalizer is applied, followed by an RBF-kernel SVM classifier, representing a classical separated source-channel coding approach.
\end{enumerate}

\subsubsection{Evaluation Metrics}

We evaluate performance using accuracy, SSIM, bandwidth efficiency, and robustness under varying SNR conditions, as summarized in Table~\ref{tab:metrics}.

\begin{table}[!t]
\centering
\caption{Evaluation Metrics}
\label{tab:metrics}
\setlength{\tabcolsep}{3pt}
\renewcommand{\arraystretch}{1.1}
\begin{tabular}{ll}
\toprule
\textbf{Metric} & \textbf{Definition} \\
\midrule

Accuracy 
& Classification correctness, $\displaystyle \text{Acc} = \frac{N_{\text{correct}}}{N_{\text{total}}}$ \\

SSIM 
& Perceptual reconstruction quality \\

Bandwidth Ratio 
& Compression efficiency, $\displaystyle R_B = \frac{L_z}{L_x}$ \\

SNR~(dB) 
& Channel quality indicator, $\displaystyle \mathrm{SNR}_{\mathrm{dB}} = 10\log_{10}\!\left(\frac{P_s}{P_n}\right)$ \\

\bottomrule
\end{tabular}
\end{table}

\begin{figure*}[!t]
    \centering

    \begin{subfigure}[b]{\columnwidth}
        \centering
        \includegraphics[width=\linewidth]{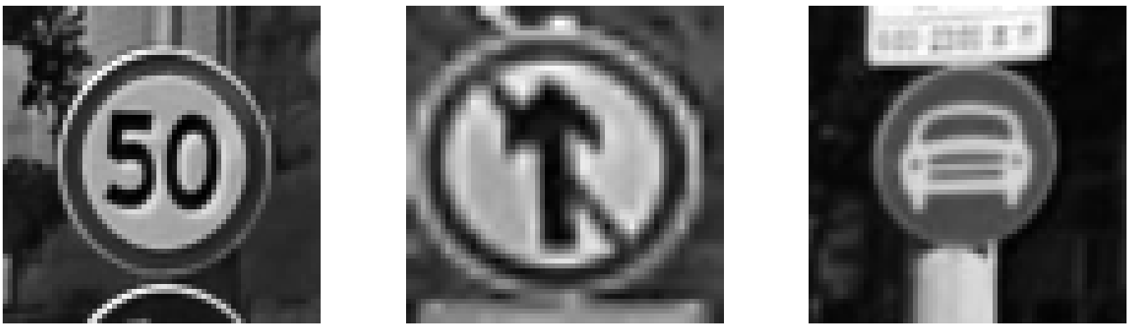}
        \caption{}
    \end{subfigure}
    \hfill
    \begin{subfigure}[b]{\columnwidth}
        \centering
        \includegraphics[width=\linewidth]{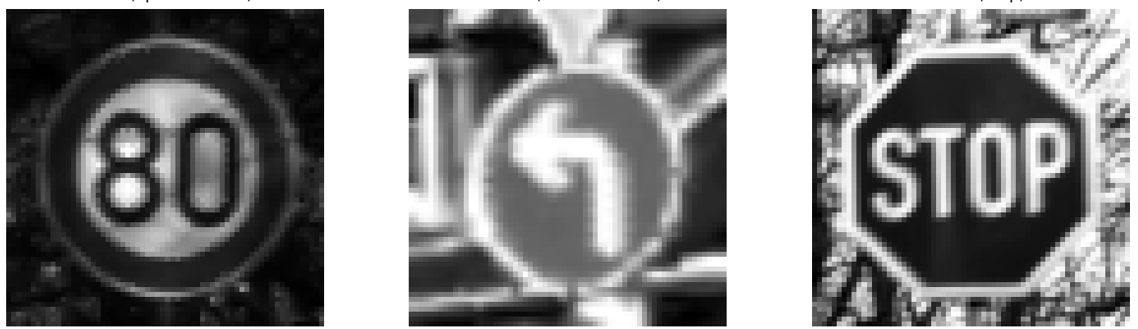}
        \caption{}
    \end{subfigure}

    \vspace{0.5em}

    \begin{subfigure}[b]{\columnwidth}
        \centering
        \includegraphics[width=\linewidth]{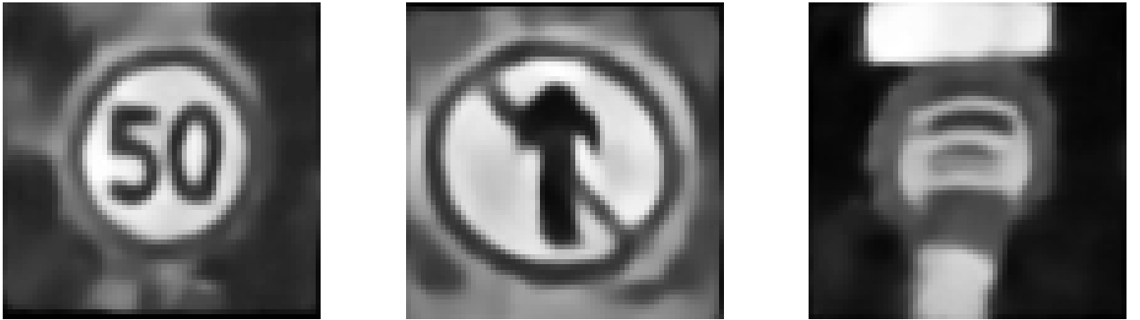}
        \caption{}
    \end{subfigure}
    \hfill
    \begin{subfigure}[b]{\columnwidth}
        \centering
        \includegraphics[width=\linewidth]{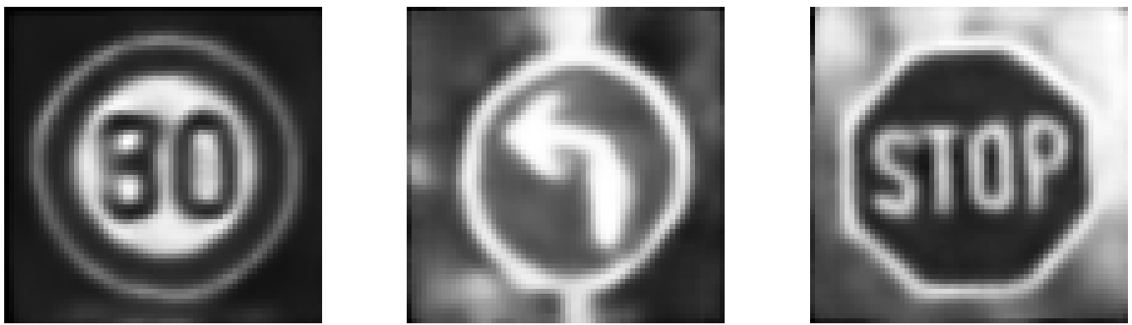}
        \caption{}
    \end{subfigure}

    \vspace{0.5em}

    \begin{subfigure}[b]{\columnwidth}
        \centering
        \includegraphics[width=\linewidth]{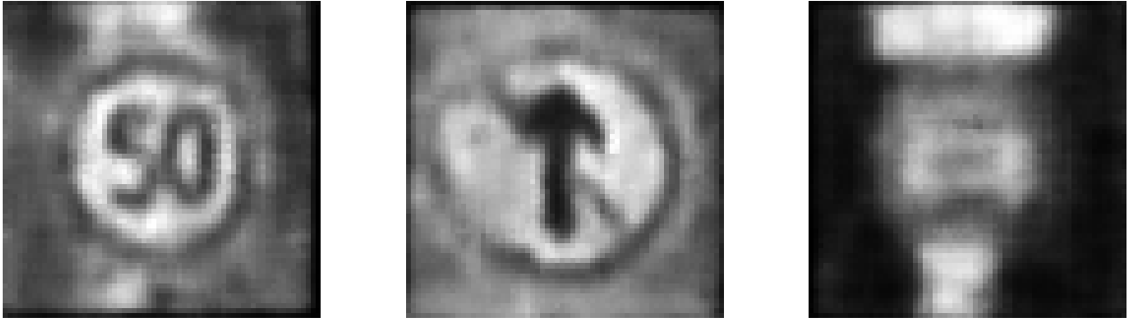}
        \caption{}
    \end{subfigure}
    \hfill
    \begin{subfigure}[b]{\columnwidth}
        \centering
        \includegraphics[width=\linewidth]{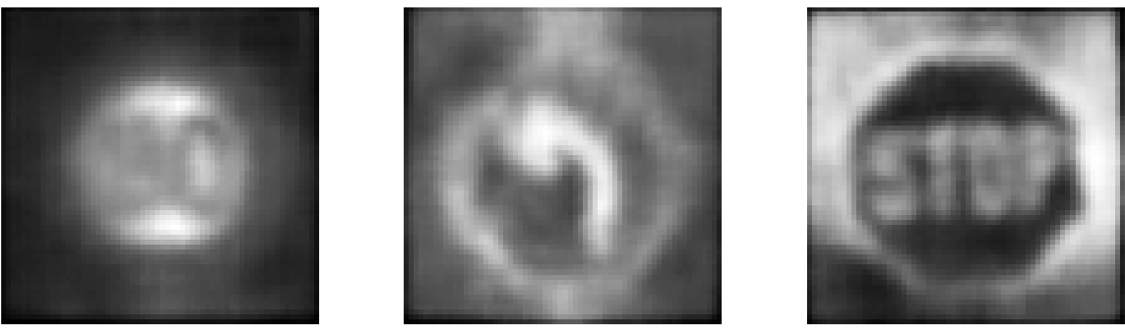}
        \caption{}
    \end{subfigure}

    \vspace{0.5em}

    \begin{subfigure}[b]{\columnwidth}
        \centering
        \includegraphics[width=\linewidth]{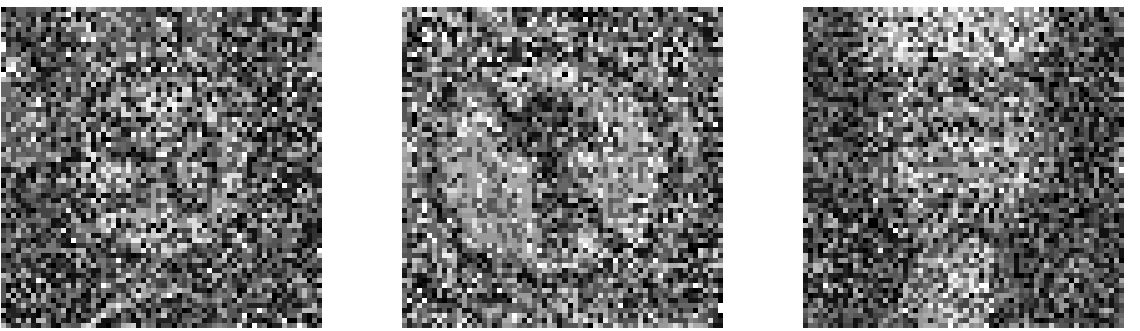}
        \caption{}
    \end{subfigure}
    \hfill
    \begin{subfigure}[b]{\columnwidth}
        \centering
        \includegraphics[width=\linewidth]{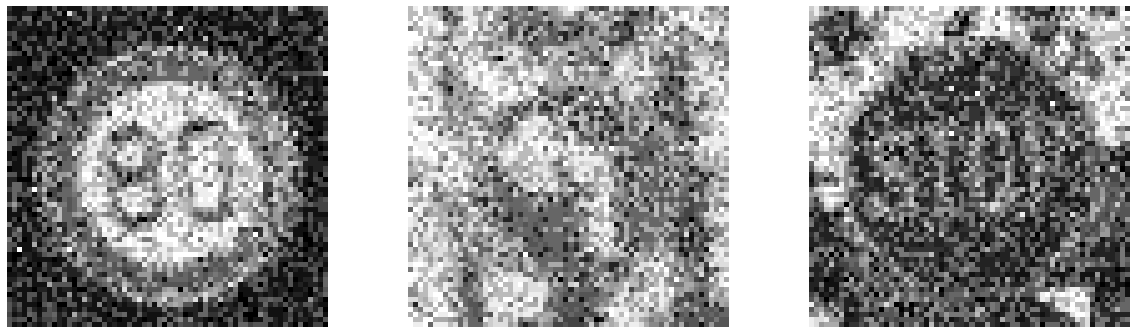}
        \caption{}
    \end{subfigure}

    \caption{Visual comparison of traffic sign reconstruction performance: (a) Chinese original, (b) German original, (c) Chinese reconstructed using VAE, (d) German reconstructed using VAE, (e) Chinese reconstructed using AE, (f) German reconstructed using AE, (g) Chinese reconstructed using QAM, and (h) German reconstructed using QAM.}
    
    \label{fig:all_comparison}
    \vspace{-6mm}
\end{figure*}

\vspace{-1.5mm}
\section{Results and Discussion}
\label{sec:results}

\subsection{Classification Accuracy vs.\ SNR}

\begin{table}[!t]
\centering
\caption{Classification Accuracy (\%) at Each Test SNR Level}
\label{tab:acc}
\setlength{\tabcolsep}{3.5pt}
\renewcommand{\arraystretch}{1.1}
\begin{tabular}{ll l *{5}{c}}
\toprule
& & & \multicolumn{5}{c}{\textbf{SNR (dB)}} \\
\cmidrule(lr){4-8}
\textbf{Dataset} & \textbf{System} & \textbf{Code} &
\textbf{$-10$} & \textbf{$0$} & \textbf{$10$} & \textbf{$20$} & \textbf{$30$} \\
\midrule

\multirow{9}{*}{\textbf{Chinese}}
  & \multirow{4}{*}{VAE (proposed)}
        & 16  & 16.23 & 55.64 & 68.74 & 69.99 & 71.47 \\
  &     & 32  & 24.49 & 64.18 & 74.54 & 73.86 & 73.41 \\
  &     & 64  & 33.20 & 67.08 & 73.58 & 75.00 & 74.43 \\
  &     & 128 & \textbf{46.81} & \textbf{73.75} & \textbf{77.90} & \textbf{78.47} & \textbf{77.56} \\
\cmidrule(lr){2-8}
  & \multirow{4}{*}{AE (baseline)}
        & 16  & 18.85 & 55.24 & 70.22 & 72.38 & 72.32 \\
  &     & 32  & 22.44 & 62.07 & 72.55 & 74.66 & 75.06 \\
  &     & 64  & 30.13 & 64.35 & 72.38 & 73.41 & 73.92 \\
  &     & 128 & 39.81 & 69.31 & 75.28 & 76.94 & 75.40 \\
\cmidrule(lr){2-8}
  & QAM-16\,+\,SVM & -- & 2.28 & 17.37 & 30.81 & 32.40 & 32.86 \\

\midrule

\multirow{9}{*}{\textbf{German}}
  & \multirow{4}{*}{VAE (proposed)}
        & 16  & 22.72 & 77.42 & 95.32 & 96.04 & 96.13 \\
  &     & 32  & 34.61 & 86.78 & 95.79 & 96.53 & 96.09 \\
  &     & 64  & 52.62 & 92.05 & 96.26 & 96.52 & 96.39 \\
  &     & 128 & \textbf{74.22} & \textbf{95.46} & \textbf{96.56} & \textbf{96.58} & \textbf{96.68} \\
\cmidrule(lr){2-8}
  & \multirow{4}{*}{AE (baseline)}
        & 16  & 20.01 & 76.77 & 92.35 & 93.35 & 93.15 \\
  &     & 32  & 27.81 & 87.06 & 93.27 & 93.38 & 93.43 \\
  &     & 64  & 35.49 & 93.16 & 95.15 & 95.40 & 95.36 \\
  &     & 128 & 49.32 & 94.24 & 96.36 & 96.48 & 96.58 \\
\cmidrule(lr){2-8}
  & QAM-16\,+\,SVM & -- & 3.03 & 15.66 & 27.03 & 25.49 & 25.03 \\

\bottomrule
\multicolumn{8}{l}{\footnotesize Bold denotes best result per dataset and SNR level.}
\end{tabular}
\vspace{-1 mm}
\end{table}


Table~\ref{tab:acc} shows the classification accuracy for both the German and Chinese datasets at different SNR levels. Across all noise conditions, the proposed 4$\times$VAE consistently outperforms the AE baseline in all noise conditions, and the performance gap increases in the low-SNR regimes. In the German dataset, the proposed method achieves an accuracy ranging from 74.22\% to 96.68\% across SNR values from -10 dB to 30 dB. Similar trends are observed for the Chinese dataset, where the proposed method achieves accuracies ranging from 46.81\% to 77.56\% over SNR values between -10 dB and 30 dB. The proposed method, VAE, maintains more robustness under severe channel noise. At moderate and high SNR levels (10-30 dB), both AE and VAE approaches converge; however, the proposed VAE still achieves marginal gains, indicating improved latent robustness and better feature preservation under channel perturbations. The classical QAM baseline suffers from severe degradation, confirming the advantage of learned joint source-channel representations.

\subsection{Reconstruction Quality (SSIM) vs.\ SNR}

\begin{table}[t]
\centering
\caption{Structural Similarity at Each Test SNR Level}
\label{tab:ssim}
\setlength{\tabcolsep}{3.5pt}
\renewcommand{\arraystretch}{1.1}
\begin{tabular}{ll l *{5}{c}}
\toprule
& & & \multicolumn{5}{c}{\textbf{SNR (dB)}} \\
\cmidrule(lr){4-8}
\textbf{Dataset} & \textbf{System} & \textbf{Code} &
\textbf{$-10$} & \textbf{$0$} & \textbf{$10$} & \textbf{$20$} & \textbf{$30$} \\
\midrule

\multirow{9}{*}{\textbf{Chinese}}
  & \multirow{4}{*}{VAE (proposed)}
        & 16  & 0.319 & 0.451 & 0.508 & 0.516 & 0.516 \\
  &     & 32  & 0.333 & 0.506 & 0.569 & 0.576 & 0.578 \\
  &     & 64  & 0.366 & 0.568 & 0.630 & 0.638 & 0.638 \\
  &     & 128 & \textbf{0.404} & \textbf{0.630} & \textbf{0.681} & \textbf{0.687} & \textbf{0.688} \\
\cmidrule(lr){2-8}
  & \multirow{4}{*}{AE (baseline)}
        & 16  & 0.103 & 0.166 & 0.192 & 0.196 & 0.197 \\
  &     & 32  & 0.112 & 0.207 & 0.246 & 0.250 & 0.250 \\
  &     & 64  & 0.134 & 0.273 & 0.316 & 0.321 & 0.322 \\
  &     & 128 & 0.170 & 0.351 & 0.404 & 0.410 & 0.411 \\
\cmidrule(lr){2-8}
  & QAM-16\,+\,SVM & -- & 0.152 & 0.262 & 0.363 & 0.382 & 0.384 \\

\midrule
 
\multirow{9}{*}{\textbf{German}}
  & \multirow{4}{*}{VAE (proposed)}
        & 16  & 0.379 & 0.604 & 0.688 & 0.699 & 0.702 \\
  &     & 32  & 0.416 & 0.669 & 0.756 & 0.766 & 0.767 \\
  &     & 64  & 0.426 & 0.733 & 0.814 & 0.823 & 0.824 \\
  &     & 128 & \textbf{0.474} & \textbf{0.799} & \textbf{0.865} & \textbf{0.872} & \textbf{0.872} \\
\cmidrule(lr){2-8}
  & \multirow{4}{*}{AE (baseline)}
        & 16  & 0.379 & 0.534 & 0.599 & 0.612 & 0.612 \\
  &     & 32  & 0.369 & 0.582 & 0.653 & 0.662 & 0.664 \\
  &     & 64  & 0.380 & 0.634 & 0.706 & 0.714 & 0.715 \\
  &     & 128 & 0.402 & 0.6836 & 0.754 & 0.762 & 0.763 \\
\cmidrule(lr){2-8}
  & QAM-16\,+\,SVM & -- & 0.119 & 0.227 & 0.406 & 0.488 & 0.502 \\
 
\bottomrule
\multicolumn{8}{l}{\footnotesize Bold denotes best result per dataset and SNR level.}
\end{tabular}
\vspace{-2.5mm}
\end{table}


Table~\ref{tab:ssim} shows SSIM results that reflect the perceptual quality of the reconstruction of transmitted images. The proposed VAE outperforms with higher SSIM values for all SNRs and datasets. The SSIM is 87.2\% in the German dataset and 68.8\% at 30dB in the Chinese dataset. Compared to AE and QAM, the VAE preserves structural details more effectively, especially under low SNR conditions where AE and QAM performance drops noticeably. This demonstrates that the stochastic latent representation of VAE leads to better robustness against channel noise, thereby providing better perceptual consistency for reconstructed images.



\subsection{Bandwidth Analysis}

\begin{table}[t]
\centering
\caption{Bandwidth Analysis: Compressed Latent Payload vs.\ Raw Image
         ($64{\times}64{\times}1 = 4096$\,bytes)}
\label{tab:bandwidth}
\setlength{\tabcolsep}{13pt}
\renewcommand{\arraystretch}{1.1}
\begin{tabular}{l r r r}
\toprule
\textbf{Scheme} & \textbf{Bytes} & \textbf{KBytes} & \textbf{Savings} \\
\midrule
Raw image (reference) & 4096 & 4.000 & ---   \\
\midrule
VAE / AE, Code\,=\,16  &  75  & 0.073 & 98.17\% \\
VAE / AE, Code\,=\,32  & 139  & 0.136 & 96.61\% \\
VAE / AE, Code\,=\,64  & 267  & 0.261 & 93.48\% \\
VAE / AE, Code\,=\,128 & 523  & 0.511 & 87.23\% \\
\bottomrule
\end{tabular}
\end{table}


Table~\ref{tab:bandwidth} highlights that both AE and VAE can achieve a significant bandwidth reduction, with a maximum of 98.17\% saving compared to raw image transmission. But efficiency in compression does not guarantee reliable performance. The proposed VAE is consistently more robust to keep semantic and structural information in the noisy channel conditions, even with similar compression ratios. On the other hand, the AE suffers a significant degradation in reconstruction quality and classification accuracy. This emphasizes that bandwidth efficiency alone is not enough, and that a strong latent representation of the VAE is essential to reliable performance in a noisy channel environment.

\subsection{Qualitative Reconstruction}
Fig.~\ref{fig:all_comparison} shows behaviors beyond what SSIM scores can capture. At $10$\,dB and above, VAE outputs maintain stroke contours, numeric annotations, and boundary geometry for clear sign identification. In contrast, the AE, despite having the same architecture, produces softer reconstructions, flattening interior details and blurring boundaries. QAM-16 reconstructions are severely degraded, primarily showing only coarse outlines due to channel noise. This issue worsens at $0$\,dB and $-10$\,dB, with VAE reconstructions maintaining structural coherence while AE outputs fall into low-contrast patches and QAM-16 results become nearly unrecognizable. The difference arises from VAE's probabilistic latent structure, allowing for better adaptation to noise compared to the deterministic AE and non-adaptive QAM-16.

\vspace{-2.5 mm}
\section{Conclusion}
\label{sec:conclusion}

In this paper, we present a multi-task semantic communication framework utilizing a VAE with a composite perceptual reconstruction loss. By combining $L_1$, MSE, SSIM, and image-gradient terms, our method improves reconstruction fidelity and classification robustness over the AE baseline and QAM transmission pipeline. The decoder refinement branch enhances perceptual quality, while a deterministic inference reduces variability during testing. We utilize a scaled KL-weighting mechanism to tackle the issue of posterior collapse, which helps us maintain stable training across different compression rates. Experimental results show that our VAE performs better than both the autoencoder baseline and the QAM-16 scheme, especially in low SNR conditions. We have successfully achieved a balance between classification accuracy, reconstruction quality, and bandwidth efficiency. Our results demonstrate that probabilistic semantic encoding is a feasible and effective strategy for bandwidth-limited satellite-assisted vehicular networks, opening a principled path towards intelligent, task-aware communication in next-generation autonomous driving infrastructure.


\bibliographystyle{ieeetr}
\bibliographystyle{unsrtnat}

\end{document}